\documentclass[sigconf]{acmart}
\usepackage{multirow}
\usepackage{subfigure}
\usepackage{hyperref}
\usepackage{balance}

\usepackage{wrapfig}

\AtBeginDocument{%
  }

\copyrightyear{2024}
\acmYear{2024}
\setcopyright{rightsretained}
\acmConference[CIKM '24]{Proceedings of the 33rd ACM International Conference on Information and Knowledge Management}{October 21--25, 2024}{Boise, ID, USA}
\acmBooktitle{Proceedings of the 33rd ACM International Conference on Information and Knowledge Management (CIKM '24), October 21--25, 2024, Boise, ID, USA}
\acmDOI{10.1145/3627673.3679702}
\acmISBN{979-8-4007-0436-9/24/10}
\acmDOI{10.1145/3627673.3679702}


\begin{document}

\title{CYCLE: Cross-Year Contrastive Learning in Entity-Linking}

\author{Pengyu Zhang}
\orcid{0000-0001-5111-4487}
\affiliation{%
  \institution{University of Amsterdam}
  \city{Amsterdam}
  \country{The Netherlands}}
\email{p.zhang@uva.nl}

\author{Congfeng Cao}
\orcid{0000-0001-9011-3807}
\affiliation{%
  \institution{University of Amsterdam}
  \city{Amsterdam}
  \country{The Netherlands}}

\author{Klim Zaporojets}
\orcid{0000-0003-4988-978X}
\affiliation{%
  \institution{Aarhus University}
  \city{Aarhus}
  \country{Denmark}}

\author{Paul Groth}
\orcid{0000-0003-0183-6910}
\affiliation{%
  \institution{University of Amsterdam}
  \city{Amsterdam}
  \country{The Netherlands}}

\renewcommand{\shortauthors}{Pengyu Zhang, Congfeng Cao, Klim Zaporojets, and Paul Groth}
\newcommand{\ourdataset}{GCL-TempEL}

\begin{abstract}
Knowledge graphs constantly evolve with new entities emerging, existing definitions being revised, and entity relationships changing. These changes lead to temporal degradation in entity linking models, characterized as a decline in model performance over time. To address this issue, we propose leveraging graph relationships to aggregate information from neighboring entities across different time periods. This approach enhances the ability to distinguish similar entities over time, thereby minimizing the impact of temporal degradation. We introduce \textbf{CYCLE}: \textbf{C}ross-\textbf{Y}ear \textbf{C}ontrastive \textbf{L}earning for \textbf{E}ntity-Linking. This model employs a novel graph contrastive learning method to tackle temporal performance degradation in entity linking tasks. Our contrastive learning method treats newly added graph relationships as \textit{positive} samples and newly removed ones as \textit{negative} samples. This approach helps our model effectively prevent temporal degradation, achieving a 13.90\% performance improvement over the state-of-the-art from 2023 when the time gap is one year, and a 17.79\% improvement as the gap expands to three years. Further analysis shows that CYCLE is particularly robust for low-degree entities, which are less resistant to temporal degradation due to their sparse connectivity, making them particularly suitable for our method. The code and data are made available at \url{https://github.com/pengyu-zhang/CYCLE-Cross-Year-Contrastive-Learning-in-Entity-Linking}.
\end{abstract}



\maketitle

\section{Introduction}

Knowledge graphs (KGs) are multi-relational graphs representing a wide range of real-world entities and knowledge structured as facts \cite{10.24963/ijcai.2023/734}. In KGs, facts are represented as triples, denoted as \textit{(head entity, relation, tail entity)}. KGs like ICEWS \cite{boschee2015icews}, GDELT \cite{leetaru2013gdelt}, and YAGO3 \cite{mahdisoltani2013yago3} not only contain numerous entities and their relations but also incorporate timestamps, allowing to track the evolution of the knowledge in these KGs. Each of these timestamps is part of a quadruplet \textit{(head entity, relation, tail entity, timestamp)}, reflecting the point in time a particular relation between head and tail entities was created. Such temporal features enable KG users to track historical data trends, understand entity behavior over time, and predict future events or facts, highly relevant in domains such as medical and risk analysis systems \cite{9742265}, question-answering systems \cite{sharma2022twirgcn}, and recommendation systems \cite{bogina2023considering}. However, the continuous emergence of new entities and changes to existing ones, poses challenges in adapting entity representations, which in turn affects the performance on tasks such as Entity Linking (EL) \cite{pan2023unifying}.

\begin{figure}[t]
\centering
\includegraphics[width=\linewidth]{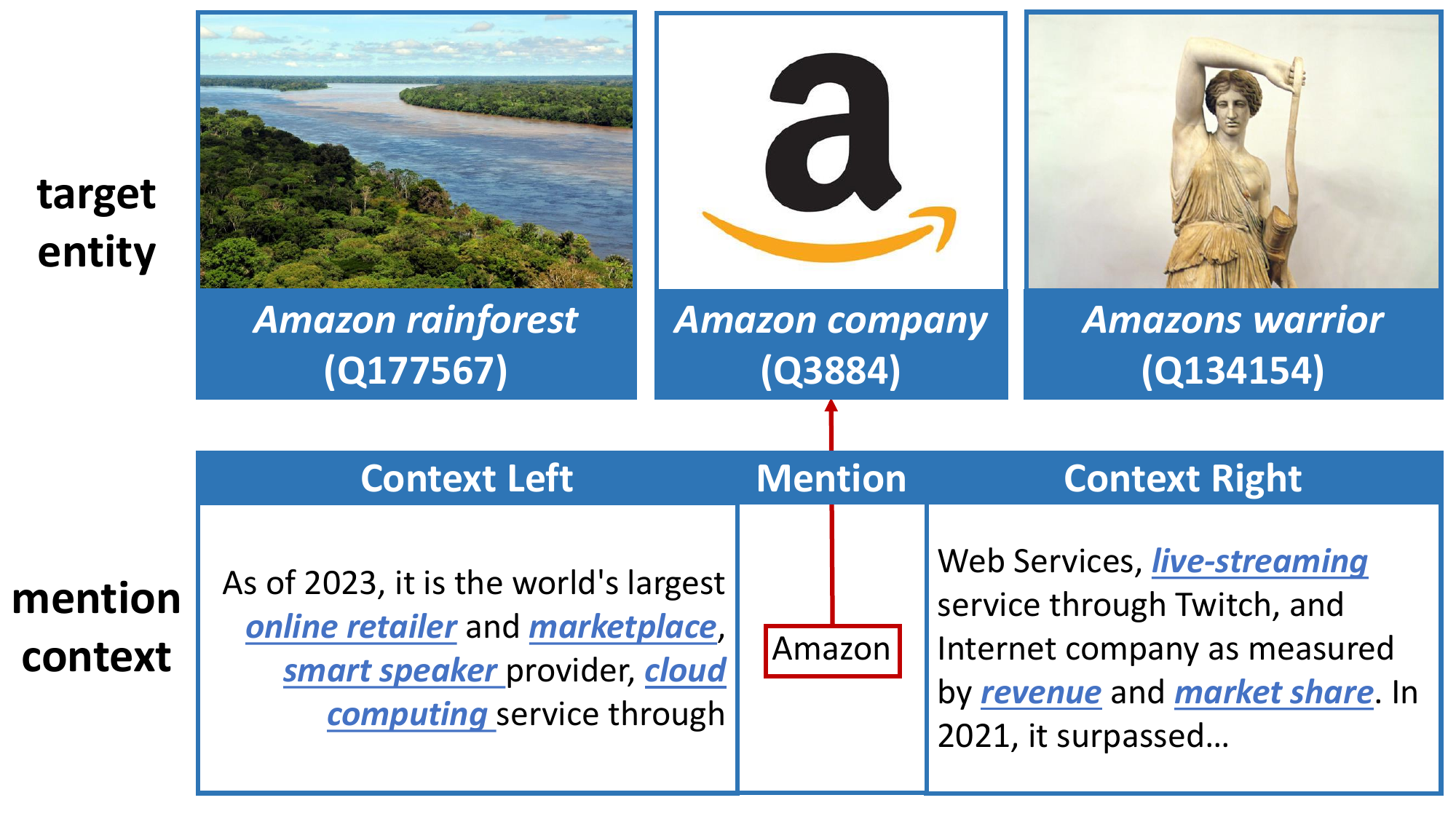}
\caption{Entity linking (EL) connects text mentions to specific entities in a KG. In the example of this figure, the mention `Amazon' could refer to \textit{Amazon rainforest}, \textit{Amazon company}, or mythical \textit{Amazons warriors} entities. EL disambiguates this mention considering the surrounding \textit{mention context} to pinpoint the correct entity: \textit{Amazon company}.}\label{figure1}
\Description{figure1}
\end{figure}

\begin{figure*}[ht]
\centering
\includegraphics[width=0.9\linewidth]{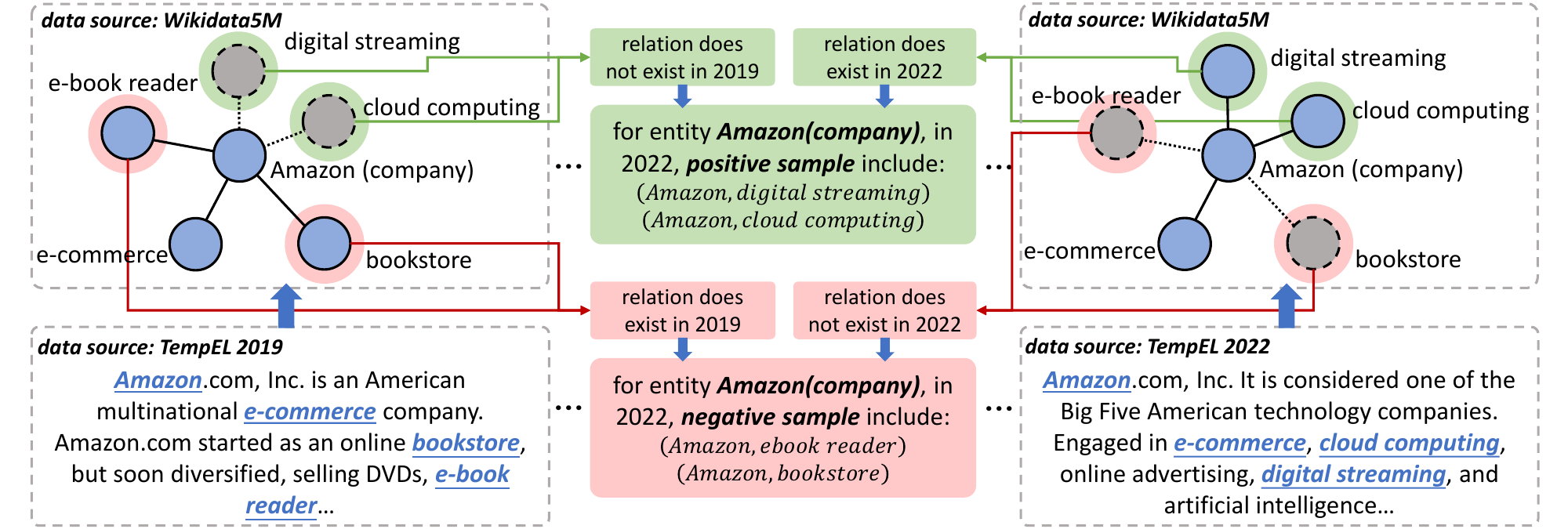}
\caption{This is an example of learning the embedding of the entity \textit{Amazon (company)} in 2022. Gray nodes represent entities that do not exist at a specific point in time. A (\textcolor[rgb]{0.663, 0.82, 0.557}{green}) or (\textcolor[rgb]{1, 0.671, 0.671}{red}) circle represents the use of a node as a \textit{positive} or \textit{negative} sample, respectively. In 2019, \textit{Amazon (company)} had three neighbors: \textit{e-commerce}, \textit{e-book reader}, and \textit{bookstore}. By 2022, while \textit{e-commerce} continued as a neighbor, \textit{e-book reader} and \textit{bookstore} (negative samples) were replaced by \textit{digital streaming} and \textit{cloud computing} (positive samples). This shift indicates a change in \textit{Amazon (company)}'s focus.
}\label{figure2}
\Description{figure2}
\end{figure*}

EL involves mapping mentions in text to entities in a KG. In the example in Figure~\ref{figure1}, EL maps the mention `Amazon' in the text, with its correct entity \textit{Amazon (company)} in the KG. The term `Amazon' could refer to \textit{Amazon (company)}, known for e-commerce, or \textit{Amazon rainforest}. The correct entity is decided by analyzing the content of the context. If the context includes \textit{cloud computing}, it likely refers to \textit{Amazon (company)}. Beyond the challenge of disambiguating the correct entity given the mention and its context, this task becomes even more difficult in the face of the constant addition of new entities in a KG and the evolving meanings of existing entities (continual entities). This change leads to temporal degradation - a decline in model performance as the KG moves from the state where the EL model was initially trained.

In this paper, we develop a new model for EL that deals with temporal change. For instance, as shown in Figure~\ref{figure2}, in 2019, the primary sources of income for \textit{Amazon (company)} were \textit{e-book reader} and \textit{bookstore}. By 2022, however, \textit{Amazon (company)} had established new connections, and recently added entities like \textit{digital streaming} and \textit{cloud computing} had become major contributors to its income. Concentrating on these new connections allows for a deeper understanding of the current state of the entity and its context. Although these changes might appear minor, they substantially affect the model's ability to represent entities accurately.
 
To address this challenge, a benchmark introduced by \cite{jang2022temporalwiki} leverages the differences between consecutive snapshots of Wikipedia and Wikidata to provide a platform for testing and training language models, enabling them to adapt to and update continually changing knowledge information. Furthermore, recent work has focused on reducing bias through graph contrastive learning \cite{wang2022uncovering}. In addition, recent studies like \cite{zhang-etal-2022-knowledge} on self-supervised biomedical EL and \cite{10.1145/3477495.3531867} on multimodal EL with contrastive learning have advanced the field by improving accuracy. However, existing methods do not leverage the temporal evolution of structured relations between entities in a KG across different years. 
To tackle this research gap, we introduce an expanded version of the TempEL \cite{zaporojets2022tempel} dataset named \ourdataset~(see Section \ref{s4}). The original TempEL dataset consists of 10 yearly snapshots, evenly distributed, from English Wikipedia entities, spanning from January 1, 2013, to January 1, 2022. Building upon this, we have incorporated the relationships between entities for each year from the KG introduced in the Wikidata5M \cite{wang2021kepler} dataset, and the changes in these relationships over time.

We hypothesize that the evolution of KG relationships across different years in our newly introduced dataset, can provide crucial information about the changes in the entities (see Figure~\ref{figure2}). To support this, we introduce \textbf{CYCLE}: \textbf{C}ross-\textbf{Y}ear \textbf{C}ontrastive \textbf{L}earning in \textbf{E}ntity-Linking. \textbf{CYCLE} is a novel approach to solve EL task in temporally evolving setting based on graph contrastive learning, leveraging the features of temporal data to construct a cross-year contrastive mechanism. Doing so ensures that similar entity representations remain distinct over time. Our contrastive learning approach uses KG relationships to obtain structurally enhanced entity representations. We define such \textit{positive} and \textit{negative} samples as follows:

\begin{enumerate}
\item [1] \textit{Positive samples}: sampled from a pool of newly added relationships between the target entity and its neighbors.
\item [2] \textit{Negative samples}: sampled from a pool of newly deleted relationships between the target entity and its neighbors.
\end{enumerate}

Our experimental results demonstrate that our approach can be particularly advantageous in updating representations of less frequent, long tail, entities. Such entities have lower degree connectivity in a KG and, as a result, are inherently more vulnerable to the effects of temporal changes in their neighbors. Concretely, a change in the meaning of even one neighboring node can significantly impact the low-degree node's representation. This phenomenon has been also observed in other large-scale KGs such as ICEWS and GDELT \cite{wang-etal-2019-tackling, wu2022hierarchical, wang2023survey}. 

Our contributions are summarized as follows:

\begin{itemize}
\item A dataset, \ourdataset, that incorporate cross-year KG temporal tracking of entity changes. Concretely, we define \textit{positive} and \textit{negative} samples with respect to a specific temporal snapshot.  
\item CYCLE: a novel model employing graph contrastive learning to mitigate temporal degradation in EL. CYCLE enhances the temporal stability of low-degree node representations, which are more vulnerable to semantic changes due to sparse connections.
\item Experiments across three EL datasets that demonstrate substantial improvements in model performance, particularly for low-degree nodes. The experiments highlight CYCLE's effectiveness in addressing structural vulnerabilities in KGs.
\end{itemize}

\section{Related Work}

\subsection{Entity Linking}

Generally, the Entity Linking (EL) task is categorized into three main phases: mention detection, candidate generation, and candidate ranking. Recent studies have developed end-to-end models that integrate all three phases into a single process \cite{ravi2021cholan, nedelchev2020end, broscheit2020investigating}. Specifically, the CHOLAN model \cite{ravi2021cholan} proposed two transformer-based models integrated sequentially to tackle the EL task. The first transformer identifies entity mentions in the text, while the second assigns each mention to a predefined candidate, enhanced by contextual data from the sentence and Wikipedia. As the field evolved, addressing the linking of mentions to previously unseen entities - a scenario termed zero-shot EL - remained a significant challenge \cite{10.1145/3460210.3493549}. To tackle this, \cite{wu2019zero} introduce BLINK, a highly effective two-stage BERT-based architecture for zero-shot entity linking. This model and others \cite{10.1145/3460210.3493549, bhargav-etal-2022-zero, yao-etal-2020-zero} rely on traditional candidate retrieval methods and employ a cross-encoder for candidate ranking. Building on the BLINK model, KG-ZESHEL \cite{10.1145/3460210.3493549} aims to combine graph vectors with textual content to address the zero-shot problem. Their method enhances the model's ability to clarify ambiguities and improve EL accuracy.

However, the above work did not capture the impact of changes in entity relationships on future predictions when faced with dynamic KGs. To tackle this issue, our study introduces a cross-year contrastive mechanism to capture KG relationship changes across the years, thereby improving the performance on EL task in a temporally evolving setting.

\subsection{Temporal Knowledge Graphs}

Temporal Knowledge Graphs (TKGs), capturing the dynamic evolution of entities and their relationships over time, are gaining increased attention \cite{wang2023survey}. While most of the existing graph datasets are designed for static graphs, TKGs stand out for their ability to chronicle the continuous evolution of both entities and relations. 

The application of TKGs spans various sectors, each benefiting from its capacity to track changes over time. For e-commerce, TKGs enhance understanding of consumer behavior patterns over time \cite{10.1145/3531267}. Similarly, the Internet of Things (IoT) provides a dynamic framework for interpreting evolving data from interconnected devices \cite{LI2023100441}. Healthcare applications of TKGs are particularly noteworthy, as they aid in tracking the progression of diseases and patient health trends \cite{9742265}. In industrial settings, TKGs play a pivotal role in monitoring and predicting machinery's lifecycle and maintenance needs. Recent advances in large language models (LLMs) have further expanded the potential of TKGs, particularly in forecasting applications \cite{liao2023gentkg}. These architectures offer new ways to comprehend structured temporal data, potentially revolutionizing traditional embedding-based and rule-based applications using TKGs. Moreover, integrating temporal information into KG embedding has significantly improved model performance, underscoring the importance of time-aware approaches in knowledge representation \cite{li-etal-2023-teast, dai2022wasserstein}.

However, these studies overlook the challenges of low-degree nodes with very sparse connectivity, which can significantly impact the accuracy and robustness of temporal predictions. To address this, our model employs graph contrastive learning on positive and negative samples (see Figure~\ref{figure2}) to enhance the neighbor information for low-degree nodes, thereby improving their representation in the graph.

\subsection{Graph Contrastive Learning}

Recent Graph Contrastive Learning (GCL) advancements highlight its growing significance in graph representation learning. PyGCL \cite{zhu2021empirical} emphasizes the importance of design elements like augmentation functions and contrasting modes. GraphCL \cite{you2021graph} stands out for its ability to learn robust representations from unlabeled graphs, though its effectiveness relies on specific data augmentation strategies. POT-GCL \cite{yu2023provable} addresses the need to maximize similarity between positive node pairs and minimize it for negative pairs, while recognizing unresolved issues due to complex graph structures. In contrast, SGCL \cite{sun2023rethinking} demonstrates the low impact of negative samples for achieving top performance. Finally, EdgePruner \cite{kato2023edgepruner} exposes GCL's vulnerability to poisoning attacks, suggesting a need for better defense mechanisms in graph learning models. These studies mark a significant evolution in GCL, indicating its potential and challenges in graph representation learning.

Our work expands the application of contrastive learning techniques within KGs to include temporal evolution. Specifically, we have developed a new method that considers the impact of both newly added and removed node relationships at each timestamp.

\section{Task Formulation and Definition}

\textbf{Entity Linking (EL).} The EL task takes a given text document $\mathbf{D}$ as input. This document is represented as a list of tokens $\left[w_1, \ldots, w_r\right]$, where $r$ indicates the length of the document. Each document contains a list of entity mentions $\mathbf{M}_{\mathbf{D}} = \left[m_1, \ldots, m_n\right]$, where each mention $m_{i}$ corresponds to a span of continuous tokens in $\mathbf{D}$, represented as $m_i=\mathbf{D}\left[x, y\right]$. An EL model subsequently yields a list of mention-entity pairs $\left\{\left(m_i, e_i\right)\right\}_{i \in[1, n]}$. Every entity $e_i$ is mapped to the corresponding entity in a Knowledge Base, such as Wikipedia. It is assumed that both the title and description of these entities are available, a standard premise in the EL task \cite{logeswaran-etal-2019-zero}.

\textbf{Graph.} A graph is defined as $G=(V, E)$, where $V$ is the set of $N$ nodes (i.e., entities) $\left\{v_1, v_2, \cdots, v_N\right\}$. $E$ is the set of $M$ edges (i.e., relationships) represented as $\left\{e_1, e_2, \cdots, e_M\right\}$,  where each $e_i$ is a pair of nodes from $V$, such as $e_{i}=(v_{a}, v_{b})$.

\section{Dataset Construction}\label{s4}

We extend the TempEL\footnote{https://cloud.ilabt.imec.be/index.php/s/RinXy8NgqdW58RW} dataset, a benchmark for temporal Entity Linking (EL), with Wikidata5M\footnote{https://deepgraphlearning.github.io/project/wikidata5m}. 
The resulting~\ourdataset~dataset comprises two text-based components inherited from TempEL, namely \textbf{entity description} and \textbf{mention context}, extended with three graph-based components: \textbf{relation graph}, \textbf{feature graph}, and \textbf{feature matrix}. Additionally, it includes both positive and negative samples.
Both the relation and feature graphs depict yearly entity relationships. The construction process is illustrated in Figure~\ref{figure3}, and it introduces the following components:

\begin{figure}[th]
\centering
\includegraphics[width=0.9\linewidth]{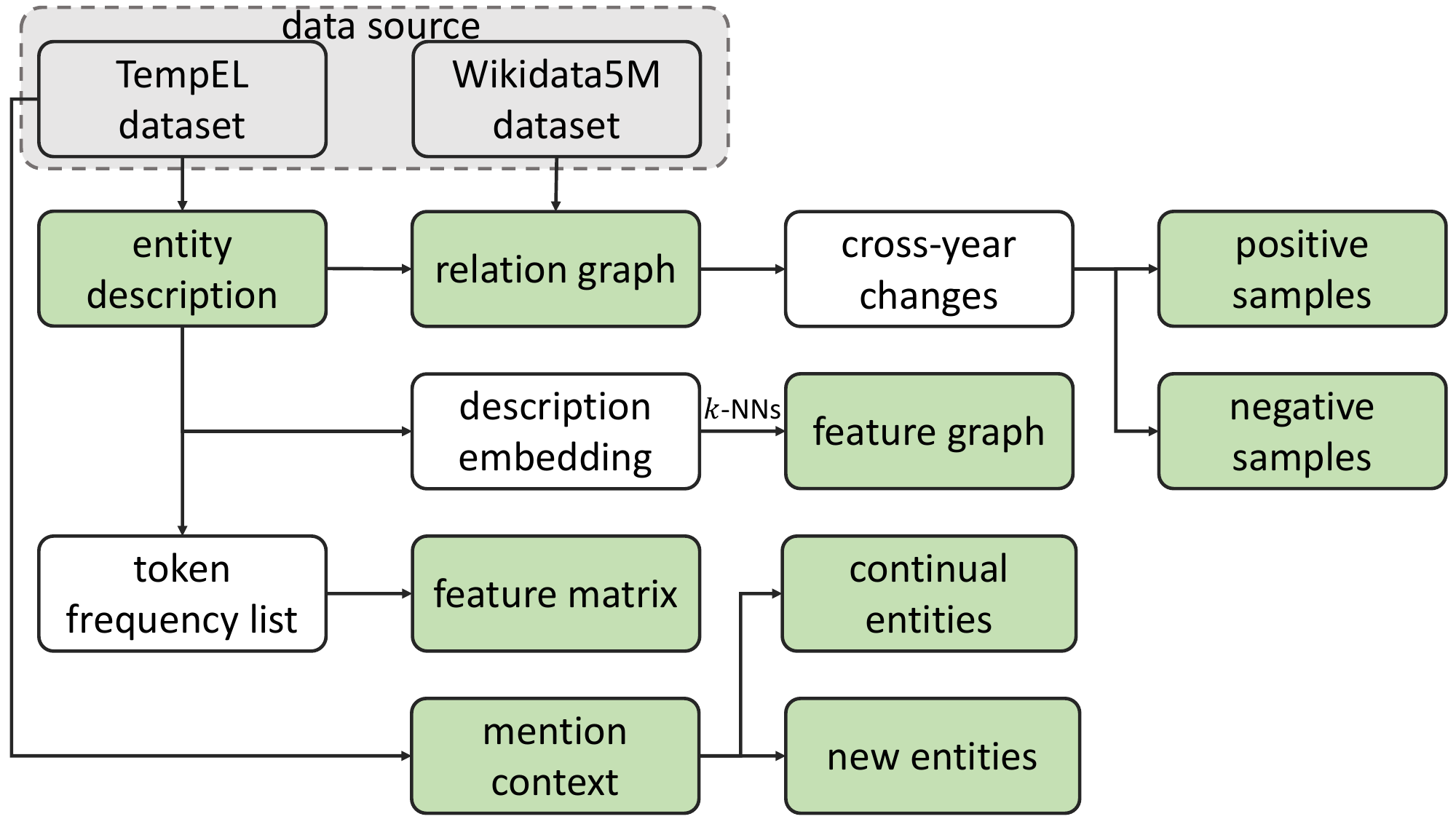}
\caption{The dataset construction process. We use Wikidata5M to extend TempEL with \textit{strutured graph} representations. For each year, we identify \textit{newly added} and \textit{removed} edges for a target entity. Furthermore, we extract \textit{feature graph} and \textit{feature matrix} based on the textual description of the target entity. The green section represents the input to our model.}\label{figure3}
\Description{figure3}
\end{figure}

\textbf{Entity descriptions and mention contexts.} First, we categorized each year of data from the TempEL dataset into \textit{entity descriptions} and \textit{mention context} parts based on the year. The \textit{entity description} comprises the title, text, document ID, and the unique ID of the entity (its QID). The \textit{mention context} consists of the text surrounding the entity mention (to the left and to the right), the mention itself, target entity as label, QID, and category. 
Furthermore, we categorized the mentions into two groups: those linked to \textit{continual entities}, which exist across all the years in \ourdataset, and those linked to \textit{new entities}, which are created in a specific year and do not exist in previous years.

\textbf{Relation graph.} We create a \textit{relation graph} based on the Knowledge Graph (KG) relationships in the Wikidata5M dataset and the entity IDs in the TempEL dataset. We matched the entities involved in the relationships in Wikidata5M with the ones that exist in the TempEL dataset. 
Concretely, we keep a relationship if both entities are involved in a relationship in the Wikidata5M data and are also present in TempEL. The relation graph is an \( n \times n \) adjacency matrix, where \( n \) represents the total number of entities in the~\ourdataset~dataset. Each row indicates whether an entity has a connection with another entity. Concretely, the adjacency matrix is made up of 0s and 1s. If entity \( i \) and entity \( j \) are connected, the value in the \( i^{th} \) row and \( j^{th} \) column of the matrix is 1; otherwise, it is 0. 
After we construct the relation graph, 
we use the differences in node 
relationships across the years (\textit{cross-year changes} in Figure \ref{figure3}) to construct \textit{positive} and \textit{negative} samples, which are defined as follows:

\begin{enumerate}
\item [1] \textit{Positive samples}: sampled from a pool of newly added relationships between the target entity and its neighbors.
\item [2] \textit{Negative samples}: sampled from a pool of newly deleted relationships between the target entity and its neighbors.
\end{enumerate}

\textbf{Feature graph.} KGs are inherently incomplete and sparse \cite{pavlovic2022expressive}.
As a result, we consider that \textit{relation graph} 
derived from Wikidata5M KG above does not contain all the possible relations between entities, which limits its expressiveness. To address this, we introduce \textit{feature graph} which extends the edges in the \textit{relation graph} with additional edges given by $k$-nearest neighbors ($k$-NN) with other entities. In order to create this graph, we use pre-trained bert-base-uncased model to embed the textual description of each of the entities. We use the resulting \textit{description embeddings} to identify the $k$-NN entities for each of the target entities using cosine similarity. The resulting \textit{feature graph} highlights the connections between entities based on their entity descriptions.
We hypothesize that such connections will provide additional information that would allow to generate more representative entity embeddings for a given temporal snapshot. Similarly to the relation graph, the feature graph is represented as \( n \times n \) adjacency matrix, where \( n \) is the total number of entities in the dataset. 
Each row indicates whether an entity has a connection with other entities. If entity \( i \) and entity \( j \) are connected, the value in the \( i^{th} \) row and \( j^{th} \) column of the matrix is 1; otherwise, it is 0.

\textbf{Feature matrix.} Building on previous research \cite{10.1145/3447548.3467415}, we developed a \textit{Feature matrix} to derive more representative entity embeddings based on their descriptions in the dataset. This matrix not only provides a robust representation of entities but also improves the graph aggregation process, enabling the generation of more nuanced embeddings. The goal of the \textit{Feature matrix} is to represent each entity based on the tokens from entity descriptions in the dataset. After obtaining the token IDs for each entity using the pre-trained bert-base-uncased model, we filtered all token IDs based on their frequency of occurrence (see \textit{token frequency list} in Figure \ref{figure3}). We retained those token IDs that appeared between 46 and 200 times. We discarded highly frequent token IDs since these tokens, such as `is,' `an,' `the,' and other common words, do not offer meaningful differentiation among entities. Also, the less frequent token IDs were removed due to the possibility of them being meaningless noise or random codes, and including an excess of these rare tokens would make the matrix too sparse, slowing down computation. 
The resulting \textit{feature matrix} is an \( n \times m \) dimensional adjacency matrix, where \( n \) is the total number of entities in the dataset, and \( m \) is the number of retained token IDs. This matrix is composed of 1s and 0s,
if the data in the \( i^{th} \) row and \( j^{th} \) column of the matrix is 1, it indicates that entity \( i \) contains the \( j^{th} \) token.

\section{Approach}

\begin{figure*}[th]
\centering
\includegraphics[width=\linewidth]{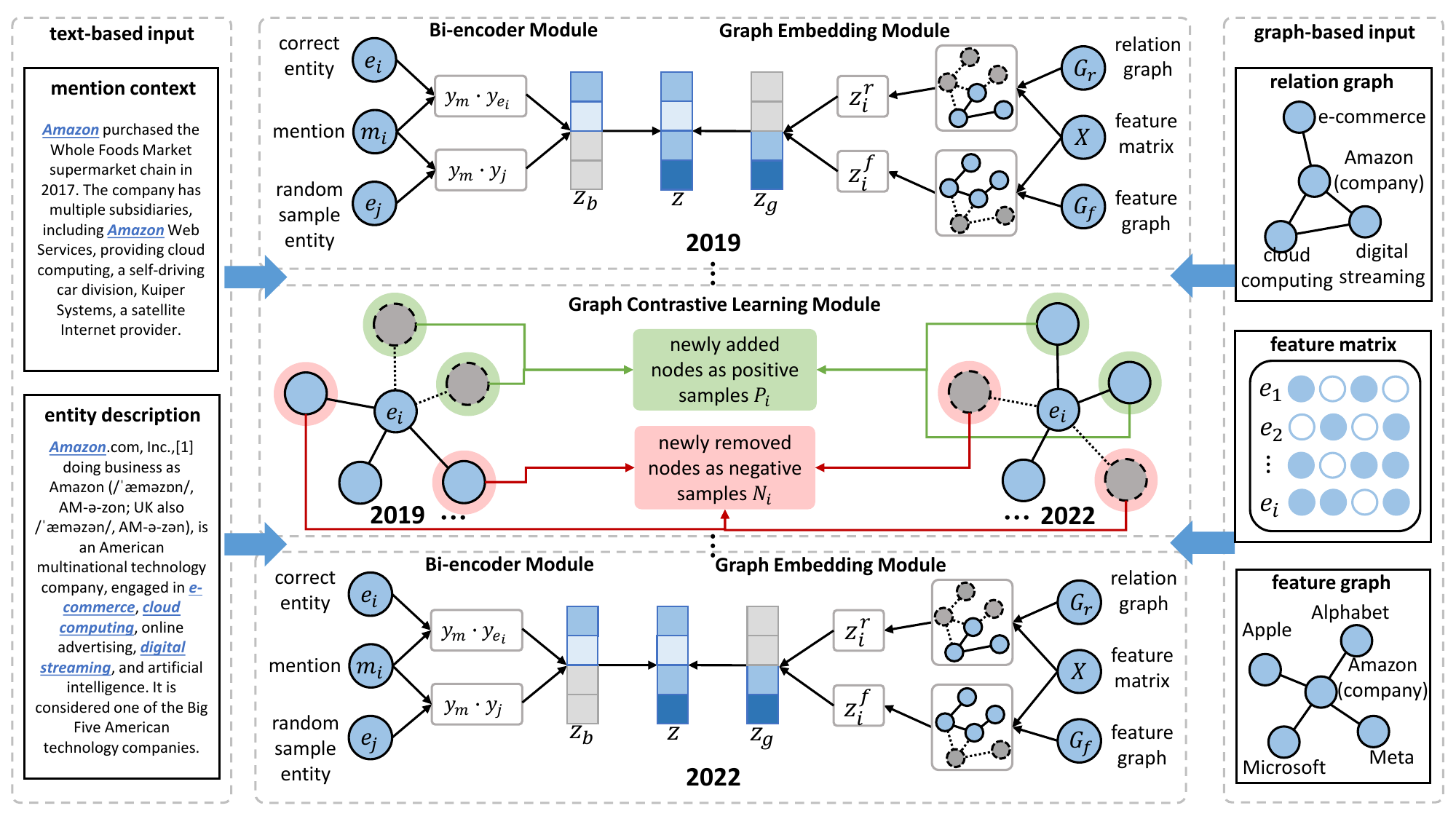}
\caption{
The proposed CYCLE architecture leverages both \textit{text-based} (left) and \textit{graph-based} (right) inputs. Additionally, it introduces a novel \textit{Graph Contrastive Learning Module} to efficiently adapt entity representations to temporal changes in the graph-based inputs.
The figure illustrates positive (\textcolor[rgb]{0.663, 0.82, 0.557}{green}) and negative (\textcolor[rgb]{1, 0.671, 0.671}{red}) samples used in this Graph Contrastive Learning Module to capture temporal changes in graph-based inputs of the year 2022 with respect to those of the year 2019 for entity $e_i$.
}\label{fig:architecture}
\Description{figure4}
\end{figure*}

Figure~\ref{fig:architecture} illustrates the framework of our model. The core idea is to exploit the relationships in \textit{graph-based input} between entities across different years in the dataset. These relationships are contained in \textit{relation graph} ($G_r$), \textit{feature graph} ($G_f$) and \textit{feature matrix} ($X$) (see Section \ref{s4} for details). When Knowledge Graph (KG) entities appear or disappear at a specific point in time, they can be used as positive and negative samples, respectively. Once these positive and negative samples are identified across different years in our~\ourdataset~dataset, we employ them in graph contrastive learning module. This module employs contrastive learning loss to efficiently adapt entity embeddings to temporal changes. Besides using the graph-based input $X$, $G_r$, $G_f$, CYCLE also leverages \textit{textual information} (see left part of Figure~\ref{fig:architecture}) composed of \textit{mention contexts} and \textit{entity descriptions}.

First, the bi-encoder module in Section \ref{s5.1} employs two separate BERT transformers to transform mention context and entity description into dense vectors $y_m$ and $y_e$. Entity candidates are scored via the dot product of these vectors. We introduce $L_e$ to maximize the correct entity's score against randomly sampled entities. Second, we input the pre-constructed relation graph $G_r$, feature graph $G_f$, feature matrix $X$, and cross-year positive and negative samples into the Graph Embedding Module in Section \ref{s5.2}. Our model encodes entities from relation and feature graphs and uses graph contrastive learning losses $L_f$ and $L_r$ across different years. 
Lastly, all the loss functions are unified for joint optimization.

\subsection{Bi-encoder Module}\label{s5.1}

\textbf{Mention Representation.} Following \cite{wu2019zero}, the mention representation $\tau_m$ is constructed from tokens of the surrounding context and the mention:
\begin{equation}
\tau_m = [\mathrm{CLS}] \operatorname{ctxt}_l\left[\mathrm{M}_s\right] \text { mention }\left[\mathrm{M}_e\right] \operatorname{ctxt}_r[\mathrm{SEP}],
\end{equation}
where ctxt$_l$, ctxt$_r$ denote tokens before and after the mention, and $\left[\mathrm{M}_s\right]$, $\left[\mathrm{M}_e\right]$ tag the mention. The input's maximum length is set to 128, consistent with the baseline.

\textbf{Entity Representation.} The representation $\tau_e$ consists of tokens of the entity title and its description:
\begin{equation}
\tau_e = [\mathrm{CLS}] \text { title }\left[\mathrm{ENT}\right] \text { description }\left[\mathrm{SEP}\right],
\end{equation}
where $\left[\mathrm{ENT}\right]$ separates the title and description.

\textbf{Encoding.} We use the bi-encoder architecture from \cite{wu2019zero} to encode descriptions into the vectors $y_e$ and $y_m$:
\begin{align}
\boldsymbol{y}_{\boldsymbol{m}} & =\operatorname{red}\left(T_1\left(\tau_m\right)\right), \\
\boldsymbol{y}_{\boldsymbol{e}} & =\operatorname{red}\left(T_2\left(\tau_e\right)\right),
\end{align}
where, $T_1$ and $T_2$ are transformers, and red(.) takes the last layer of the output of the [CLS] token to reduce the sequence of vectors into a single vector.

\textbf{Scoring.} Entity candidate scores are computed via dot-product:
\begin{equation}
s\left(m, e_i\right)=\boldsymbol{y}_{\boldsymbol{m}} \cdot \boldsymbol{y}_{\boldsymbol{e}_{\boldsymbol{i}}}.
\end{equation}

\subsection{Graph Embedding Module}\label{s5.2}

We aim to enable the learning connections between nodes from the \textit{relation graph} $G_r$. 
By inputting the relation graph $G_r$ and node features $X$, we obtain specific embeddings, denoted as $z_r$.

In the relation graph, we consider the target node $e_i$ that is connected to $s$ other nodes \{$n_1$, $n_2$,..., $n_s$\}. Thus, the set of neighbors for node $e_i$ can be defined as $N^s_i$. For node $e_i$, each neighbor contributes differently to its embedding. To effectively integrate these contributions, we employ an attention mechanism to aggregate messages from the neighbors to the target node $e_i$: 
\begin{equation}
z_i^{r}=\sigma\left(\sum_{j \in N_i^s} \alpha_{i, j} \cdot x_j\right),
\end{equation}
where $\sigma$ is a nonlinear activation, $x_j$ is the feature of node $e_j$, and $\alpha_{i,j}$ denotes the attention value of node $j$ to node $e_i$. It is calculated as follows:
\begin{equation}
\alpha_{i, j}=\frac{\exp \left(\operatorname{LeakyReLU}\left(\mathbf{a}^{\top} \cdot\left[x_i \| x_j\right]\right)\right)}{\sum_{l \in N_i} \exp \left(\operatorname{LeakyReLU}\left(\mathbf{a}^{\top} \cdot\left[x_i \| x_l\right]\right)\right)},
\end{equation}
where $\mathbf{a}$ is the attention vector and $\|$ denotes concatenate operation. Following \cite{10.1145/3447548.3467415}, 
we randomly sample a subset of neighbors in $N^s_i$ during each epoch. 
If the number of neighbors exceeds a predefined threshold, we sample neighbors as $N^s_i$. If the predefined threshold number of positive samples cannot be found, all the positive samples will be selected. In this case, the number of positive samples will be less than the predefined threshold.
This way, we ensure that every node aggregates the same amount of information from neighbors, and promote diversity of embeddings in each epoch.

When using the feature graph $G_f$ and node features $X$ as inputs, the embeddings are denoted as  \( z_{i}^{f} \).

\subsection{Cross-year Contrastive Module}

The input feature graph embedding \( z_{i}^{f} \) and relation graph embedding \( z_{i}^{r} \) are passed through a multi-layer perceptron with a single hidden layer. This process maps the features into a representation space where the contrastive loss is calculated: 
\begin{equation}
z_{i}^{fp} = W^{(2)} \sigma(W^{(1)} z_{i}^{f} + b^{(1)}) + b^{(2)},
\end{equation}
\begin{equation}
z_{i}^{rp} = W^{(2)} \sigma(W^{(1)} z_{i}^{r} + b^{(1)}) + b^{(2)},
\end{equation}
where, $\sigma$ represents exponential linear unit, a type of non-linear activation function. The parameter sets ${W^{(2)}, W^{(1)}, b^{(2)}, b^{(1)}}$ are shared across the input graphs for embedding consistency.

To define the positive $P_i^r$ and negative $N_i^r$ samples for node $e_i$ in relation graph $G_r$ at time $t_2$:

\begin{equation}
P_i^r(t_2) = \{j \mid (i, j) \notin E_{t_1} \land (i, j) \in E_{t_2}\},
\end{equation}
\begin{equation}
N_i^r(t_2) = \{j \mid (i, j) \in E_{t_1} \land (i, j) \notin E_{t_2}\},
\end{equation}
here, $E_{t_1}$ and $E_{t_2}$ denote the edge sets at times $t_1$ and $t_2$, respectively. $(i, j) \notin E_{t_1}$ indicates that there was no edge between nodes $e_i$ and $e_j$ at $t_1$, whereas $(i, j) \in E_{t_2}$ indicates that an edge between nodes $e_i$ and $e_j$ exists at time $t_2$. The symbol \(\land\) is used to denote the logical AND operation.

In a feature graph $G_f$, positive samples $P_i^f$ includes all neighboring nodes directly connected to $e_i$, while the set of negative samples $N_i^f$ includes all nodes that are not connected to $e_i$.

\subsection{Objective Function}

The objective function is calculated based on three components: two contrastive loss functions $L_{f}$ and $L_{r}$, and a task-specific EL loss function $L_e$ defined below.

\textbf{Entity Linking Loss Function $L_e$.} The objective is to train the network such that it maximizes the score of the correct entity compared to the other entities from the same batch. Formally, for each training pair $\left(m_i, e_i\right)$ within a batch of $N$ pairs, the loss $L_e$ is defined as:
\begin{equation}
{L}_e\left(m_i, e_i\right)=-s\left(m_i, e_i\right)+\log \sum_{j=1}^N \exp \left(s\left(m_i, e_j\right)\right).
\end{equation}

\textbf{Contrastive Learning Loss Functions $L_{f}$ and $L_{r}$.} 
The contrastive loss function $L_f$ and $L_r$ for a given set of positive ($P_i$) and negative ($N_i$) samples. The purpose of $L_f$ is to compute the contrastive loss where the embedding of node $e_i$ is from graph $G_f$, while the positive and negative samples are from graph $G_r$. Conversely, $L_r$ computes the contrastive loss where the embedding of node $e_i$ is from graph $G_r$, and the positive and negative samples are from graph $G_f$. This approach maximizes the advantages of contrastive learning by comparing node embeddings from different graphs and different sets of positive and negative samples, thereby capturing complex structural and semantic information from the graphs:
\begin{equation}
L_{f}=-\log \frac{\sum_{j \in P_{i}^{r}} \exp \left(\frac{\operatorname{sim}\left(z_{i}^{fp}, z_{j}^{rp}\right)}{\tau}\right)}{\sum_{k \in\left\{P_{i}^{r} \cup N_{i}^{r}\right\}} \exp \left(\frac{\operatorname{sim}\left(z_{i}^{fp}, z_{k}^{rp}\right)}{\tau}\right)},
\end{equation}

\begin{equation}
L_{r}=-\log \frac{\sum_{j \in P_{i}^{f}} \exp \left(\frac{\operatorname{sim}\left(z_{i}^{rp}, z_{j}^{fp}\right)}{\tau}\right)}{\sum_{k \in\left\{P_{i}^{f} \cup N_{i}^{f}\right\}} \exp \left(\frac{\operatorname{sim}\left(z_{i}^{rp}, z_{k}^{fp}\right)}{\tau}\right)},
\end{equation}
where the function $sim$ computes cosine similarity between two vectors, and the temperature parameter $\tau$ helps prevent the model from getting stuck in local optima during training. $P_{i}^{r}$ and $P_{i}^{f}$ are the sets of positive samples for node $e_i$, and $N_{i}^{r}$ and $N_{i}^{f}$ are the sets of negative samples for node $e_i$.

\textbf{Overall Objective Function.} 
The final objective function is a weighted sum of the individual $L_e$, $L_r$, and $L_f$ loss functions calculated above:
\begin{equation}
L=a L_{e}+b L_{f} + c L_{r},
\end{equation}
where $a$, $b$ and $c$ are weights for the three losses defined above.

\section{Evaluation}

In this section, we evaluate the performance of the proposed model across three Entity Linking (EL) datasets. The implementation of our approach is based on the original codebase BLINK\footnote{https://github.com/facebookresearch/BLINK} and HeCo\footnote{https://github.com/liun-online/HeCo} \cite{10.1145/3447548.3467415}. We selected BLINK and SpEL\footnote{https://github.com/shavarani/SpEL} \cite{shavarani-sarkar-2023-spel} because of their relevance and performance benchmarks in the field. BLINK has excellent scalability and serves as part of our model's codebase. Furthermore, SpEL is the latest state-of-the-art as of 2023. Comparison with these models highlights the influence of integrating additional structured \textit{graph-based input} and conducting graph \textit{contrastive learning} on node embeddings in CYCLE to mitigate temporal performance degradation. Experimental details can be found in \cite{zhang_2024_12790219}.

\subsection{Datasets}

Our proposed model is evaluated on three datasets which are summarized in Table~\ref{table1}. 

\begin{table}[htbp]
\centering
\caption{Summary of the used datasets' statistics.}
\label{table1}
\setlength{\tabcolsep}{1mm}{
\begin{tabular}{@{}ccccc@{}}
\toprule
                                                                   \textbf{Dataset}    & \textbf{Train} & \textbf{Validation} & \textbf{Test} & \textbf{Entities} \\ \midrule
\textbf{\begin{tabular}[c]{@{}c@{}}\ourdataset:\\ Continual Entities\end{tabular}} & 1,764          & 42,096              & 48,215        & 136,227           \\
\textbf{\begin{tabular}[c]{@{}c@{}}\ourdataset: \\ New Entities\end{tabular}}      & 1,764          & 42,096              & 48,215        & 136,227           \\
\textbf{ZESHEL}                                                                        & 49,275         & 10,000              & 10,000        & 492,321           \\
\textbf{WikiLinksNED}                                                                  & 2,188,782      & 10,000              & 10,000        & 5,455,160         \\ \bottomrule
\end{tabular}}
\end{table}

\textbf{\ourdataset}.\footnote{https://doi.org/10.5281/zenodo.12794944} This dataset covers four years, from 2019 to 2022. \ourdataset~is extended from Graph-TempEL\footnote{https://doi.org/10.5281/zenodo.12794960} and incorporates additional \textit{positive} and \textit{negative} samples with respect to specific temporal snapshot. Each year within the dataset is divided into training (1,764 samples), validation (approximately 42k samples, matching the original TempEL dataset), and testing sets (approximately 48k samples, also matching the original TempEL dataset). The number of entities remains consistent across all temporal snapshots. It is important to note that the training set includes only 1,764 samples because the original TempEL training dataset, which contains around 130,000 samples, has only 1,764 samples of \textit{new entities}. The \textit{new entities} subset, reflecting evolving trends in KGs, comprises entities with limited historical data, posing a greater challenge for accurate distinction. To ensure a balanced comparison, we also randomly selected 1,764 samples for \textit{continual entities}.

\textbf{Zero-shot Entity Linking (ZESHEL)}.\footnote{https://github.com/facebookresearch/BLINK/tree/main/examples/zeshel} This dataset covers various subjects, such as a fictional universe from a book or film series, mentions, and entities with detailed document descriptions. The train, validation, and test sets have 49K, 10K, and 10K examples, respectively. Specifically, the training set includes the following domains: `american football', `doctor who', `fallout', `final fantasy', `military', `pro wrestling', `star wars', `world of warcraft'. The validation set includes `coronation street', `muppets', `ice hockey', and `elder scrolls' domains.  Finally, the test set includes `forgotten realms', `lego', `star trek', and `Yugioh' domains. A distinctive feature of this dataset is the variation in domains between the training set and the validation and test sets. This setup effectively reflects integrating newly added entities into a Knowledge Graph (KG). The dataset features a range of 10K to 100K entity candidates per domain, with a total of 500K entities.

\textbf{WikiLinksNED}.\footnote{https://github.com/yasumasaonoe/ET4EL} This dataset was curated to address the challenges in the field of named entity disambiguation \cite{singh2012wikilinks}. Spanning a wide array of topics, from historical events to contemporary figures, the entities in this dataset are associated with detailed document descriptions. The dataset is partitioned into train, dev, and test sets with 2.1M, 10K, and 10K examples, respectively.

\subsection{Training Details}

We compare our approach to the zero-shot EL BLINK model \cite{wu2019zero}. Concretely, we reuse the same hyperparameter settings and the same bert\_uncased\_L-8\_H-512\_A-8 pre-trained model to train the bi-encoder.

\textbf{Parameter Settings.} We utilize recall@$N$ as our evaluation metric, with $N$ being 1, 2, 4, 8, 16, 32, and 64. A prediction is correct if the true answer is within the model's top $N$ predictions. The bi-encoder model is trained on the ZESHEL dataset for 5 epochs, using mention context and entity description tokens
at a learning rate 1e-05. On the~\ourdataset~dataset, training is performed for 1 epoch under similar conditions. The model undergoes annual training and testing on test sets from 2019 to 2022, with each year's model being trained and validated on that year's data before testing across other years. 

\textbf{Training Environment and Inference Time.} Software versions: Python 3.11.5; PyTorch 2.1.0\_cuda 12.1\_cudnn 8.9.2; Faiss-gpu 1.6.5; Numpy 1.26; SciPy 1.11.3; scikit-learn 1.3.2. All the experiments were run on a single A100 GPU with 40GB RAM.

\subsection{Main Results}

Table~\ref{table2} showcases the effectiveness of our model in addressing temporal degradation on the~\ourdataset~dataset. We evaluate the performance on \textit{continual} and \textit{new} entities sets. 
Each column in the table represents the years' gap between the training and testing datasets, as denoted by the digits from 0 to 3. For instance, 0 indicates that training and testing datasets come from the same year, while 3 indicates that the model was trained in 2019 and tested in 2022. The rows are divided based on various metrics: @1 to @64. `Boost' displays a comparison between our model CYCLE and SpEL model, calculated as \( \text{Boost} = \frac{\text{Our Model's Result} - \text{SpEL Model's Result}}{\text{SpEL Model's Result}} \).

\begin{table}[tbp]
\caption{
The performance of models on the~\ourdataset~dataset across varying time gaps for both \textit{new} and \textit{continual} entity sets. The best results are highlighted in bold.}
\label{table2}
\setlength{\tabcolsep}{0.5mm}{
\begin{tabular}{@{}cc|cccc|cccc@{}}
\toprule
                              &                     & \textbf{0} & \textbf{1} & \textbf{2} & \textbf{3} & \textbf{0} & \textbf{1} & \textbf{2} & \textbf{3} \\ \midrule
                              &                     & \multicolumn{4}{c|}{\textbf{Continual Entities}}  & \multicolumn{4}{c}{\textbf{New Entities}}         \\ \midrule
\multirow{4}{*}{\textbf{@1}}  & \textbf{BLINK}      & 0.177          & 0.181          & 0.182          & 0.177          & 0.132          & 0.132          & 0.132          & 0.142          \\
                              & \textbf{SpEL}       & 0.229          & 0.234          & 0.228          & 0.221          & 0.172          & 0.169          & 0.167          & 0.192          \\
                              & \textbf{CYCLE}      & \textbf{0.286} & \textbf{0.289} & \textbf{0.297} & \textbf{0.301} & \textbf{0.184} & \textbf{0.191} & \textbf{0.192} & \textbf{0.222} \\
                              & \textbf{Boost (\%)} & 25.12          & 23.78          & 30.24          & 36.26          & 6.99           & 12.86          & 14.68          & 15.28          \\ \midrule
\multirow{4}{*}{\textbf{@2}}  & \textbf{BLINK}      & 0.260          & 0.265          & 0.268          & 0.263          & 0.197          & 0.197          & 0.198          & 0.211          \\
                              & \textbf{SpEL}       & 0.320          & 0.328          & 0.327          & 0.322          & 0.239          & 0.247          & 0.258          & 0.261          \\
                              & \textbf{CYCLE}      & \textbf{0.403} & \textbf{0.409} & \textbf{0.413} & \textbf{0.413} & \textbf{0.271} & \textbf{0.282} & \textbf{0.274} & \textbf{0.331} \\
                              & \textbf{Boost (\%)} & 26.11          & 24.73          & 26.27          & 28.31          & 13.23          & 14.15          & 6.29           & 26.73          \\ \midrule
\multirow{4}{*}{\textbf{@4}}  & \textbf{BLINK}      & 0.357          & 0.364          & 0.367          & 0.362          & 0.277          & 0.277          & 0.278          & 0.294          \\
                              & \textbf{SpEL}       & 0.429          & 0.436          & 0.430          & 0.429          & 0.329          & 0.340          & 0.333          & 0.354          \\
                              & \textbf{CYCLE}      & \textbf{0.522} & \textbf{0.527} & \textbf{0.532} & \textbf{0.543} & \textbf{0.378} & \textbf{0.389} & \textbf{0.388} & \textbf{0.434} \\
                              & \textbf{Boost (\%)} & 21.65          & 21.09          & 23.73          & 26.55          & 14.80          & 14.52          & 16.31          & 22.49          \\ \midrule
\multirow{4}{*}{\textbf{@8}}  & \textbf{BLINK}      & 0.463          & 0.469          & 0.475          & 0.470          & 0.370          & 0.370          & 0.374          & 0.392          \\
                              & \textbf{SpEL}       & 0.546          & 0.544          & 0.554          & 0.548          & 0.423          & 0.440          & 0.439          & 0.472          \\
                              & \textbf{CYCLE}      & \textbf{0.633} & \textbf{0.640} & \textbf{0.647} & \textbf{0.648} & \textbf{0.487} & \textbf{0.497} & \textbf{0.491} & \textbf{0.548} \\
                              & \textbf{Boost (\%)} & 15.94          & 17.60          & 16.77          & 18.30          & 15.11          & 12.93          & 12.00          & 16.25          \\ \midrule
\multirow{4}{*}{\textbf{@16}} & \textbf{BLINK}      & 0.571          & 0.576          & 0.581          & 0.578          & 0.472          & 0.471          & 0.474          & 0.491          \\
                              & \textbf{SpEL}       & 0.645          & 0.645          & 0.656          & 0.652          & 0.539          & 0.541          & 0.554          & 0.551          \\
                              & \textbf{CYCLE}      & \textbf{0.719} & \textbf{0.719} & \textbf{0.724} & \textbf{0.715} & \textbf{0.593} & \textbf{0.602} & \textbf{0.596} & \textbf{0.644} \\
                              & \textbf{Boost (\%)} & 11.58          & 11.54          & 10.31          & 9.76           & 9.97           & 11.25          & 7.61           & 16.98          \\ \midrule
\multirow{4}{*}{\textbf{@32}} & \textbf{BLINK}      & 0.675          & 0.680          & 0.685          & 0.683          & 0.576          & 0.576          & 0.577          & 0.593          \\
                              & \textbf{SpEL}       & 0.732          & 0.739          & 0.744          & 0.741          & 0.641          & 0.646          & 0.637          & 0.673          \\
                              & \textbf{CYCLE}      & \textbf{0.811} & \textbf{0.812} & \textbf{0.817} & \textbf{0.812} & \textbf{0.689} & \textbf{0.697} & \textbf{0.688} & \textbf{0.731} \\
                              & \textbf{Boost (\%)} & 10.90          & 9.91           & 9.81           & 9.64           & 7.54           & 7.96           & 7.98           & 8.56           \\ \midrule
\multirow{4}{*}{\textbf{@64}} & \textbf{BLINK}      & 0.769          & 0.774          & 0.778          & 0.776          & 0.677          & 0.676          & 0.679          & 0.694          \\
                              & \textbf{SpEL}       & 0.820          & 0.827          & 0.825          & 0.824          & 0.732          & 0.733          & 0.739          & 0.754          \\
                              & \textbf{CYCLE}      & \textbf{0.871} & \textbf{0.872} & \textbf{0.874} & \textbf{0.878} & \textbf{0.780} & \textbf{0.784} & \textbf{0.778} & \textbf{0.811} \\
                              & \textbf{Boost (\%)} & 6.30           & 5.42           & 5.94           & 6.49           & 6.59           & 6.93           & 5.27           & 7.52           \\ \midrule
\multicolumn{2}{c|}{\textbf{Ave. Boost (\%)}}       & 16.80          & 16.29          & 17.58          & 19.33          & 10.60          & 11.51          & 10.02          & 16.26          \\ \bottomrule
\end{tabular}}
\end{table}

Table~\ref{table2} demonstrates that our model consistently outperforms the BLINK baseline as well as SpEL, with its superiority becoming increasingly clear as the temporal gap between training and testing datasets grows. Specifically, when the temporal gap is one year, and the evaluation metric is @1, our model exhibits a 23.78\% (continual entities) and 12.86\% (new entities) improvement. This improvement boosts to 30.24\% (continual entities) and 14.68\% (new entities) with a two-year gap and jumps to 36.26\% (continual entities) and 15.25\% (new entities) when the gap extends to three years. The observed performance improvement can be attributed to the baseline model's limited exposure to previously unseen new entities, resulting in a lack of samples for effective learning. In our model, graph contrastive learning enhances the distinction of each node's unique characteristics. This approach effectively enables the model to identify and accurately represent new entities, even without similar historical data. 

\begin{figure}[t]
\centering
\includegraphics[width=0.93\linewidth]{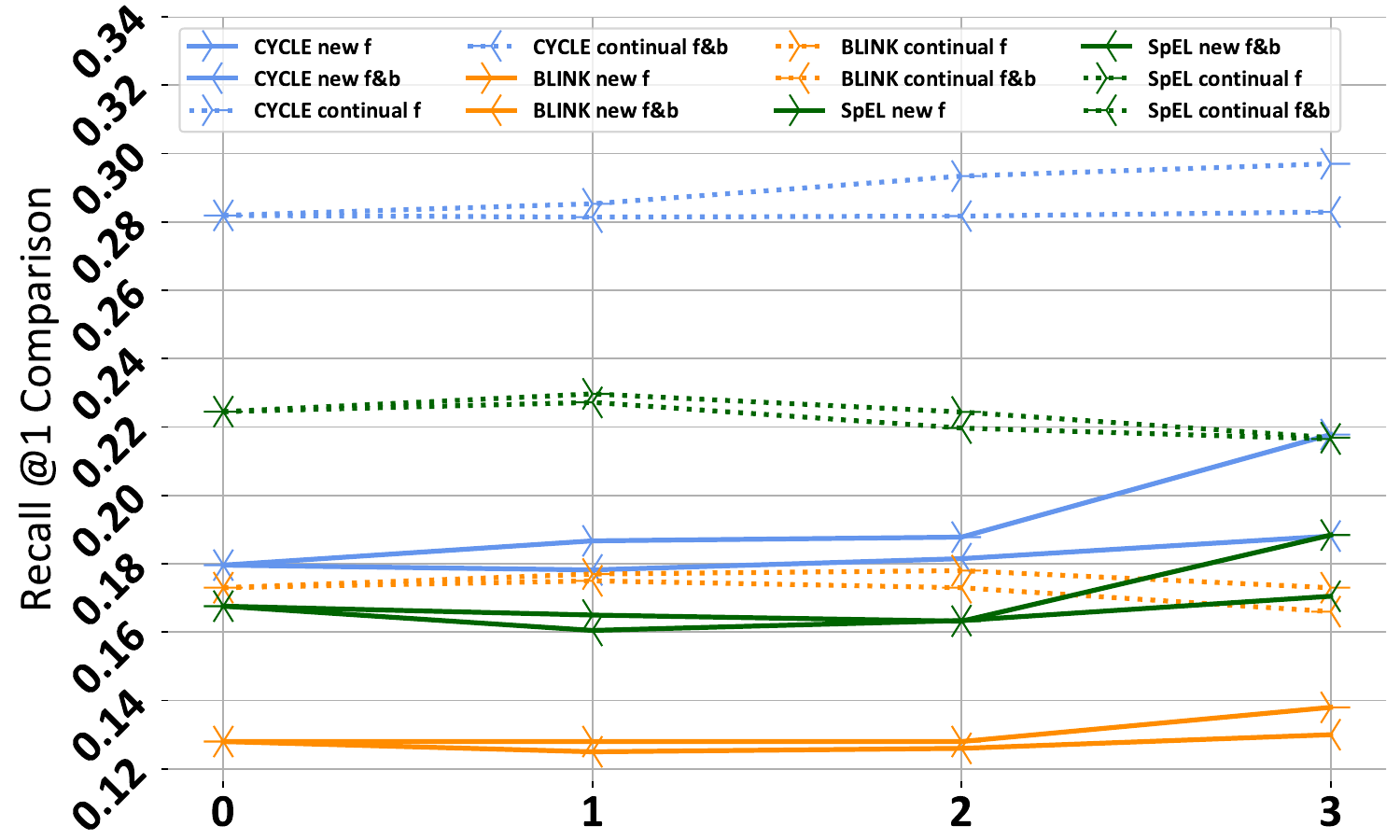}
\caption{Recall@1 performance over 3-year gaps shows CYCLE outperforming both the BLINK and the SpEL on \textit{continual} and \textit{new} entities. Performance gains increase with larger temporal gaps, highlighting the robustness of our approach.
}\label{figure5}
\Description{figure5}
\end{figure}

Figure~\ref{figure5} displays recall@1 results from the continual and new entities datasets. 
We compare our proposed model against the baseline models. The x-axis indicates the year gap between training and testing sets. Overall, our model consistently outperforms the baselines. Two different evaluation approaches are examined: 1) \textit{forward and backward (f \& b)}, referring to training on the past and testing on the future data, and vice versa, and 2) \textit{only forward (f)} where the models are only trained on the past and evaluated on the future data. For example, when the year gap is 3, the \textit{forward and backward (f \& b)} setting includes two scenarios: the model trains in 2019 and tests in 2022, and trains in 2022 and tests in 2019. However, the \textit{only forward (f)} setting only includes one scenario: the model trains in 2019 and tests in 2022. Notably, a zero-year gap implies the same years for training and testing are used, leading to equal recall values in both \textit{forward and backward (f \& b)} and \textit{only forward (f)} setting. Our model's \textit{only forward (f)} setting consistently outperforms its \textit{forward and backward (f \& b)} counterpart. We believe this demonstrates that when using previously non-existent entities as the training set, our model is more adept at capturing the evolving trends of KGs, thereby enhancing its ability to predict future developments.

\begin{figure}[t]
\centering
\includegraphics[width=0.93\linewidth]{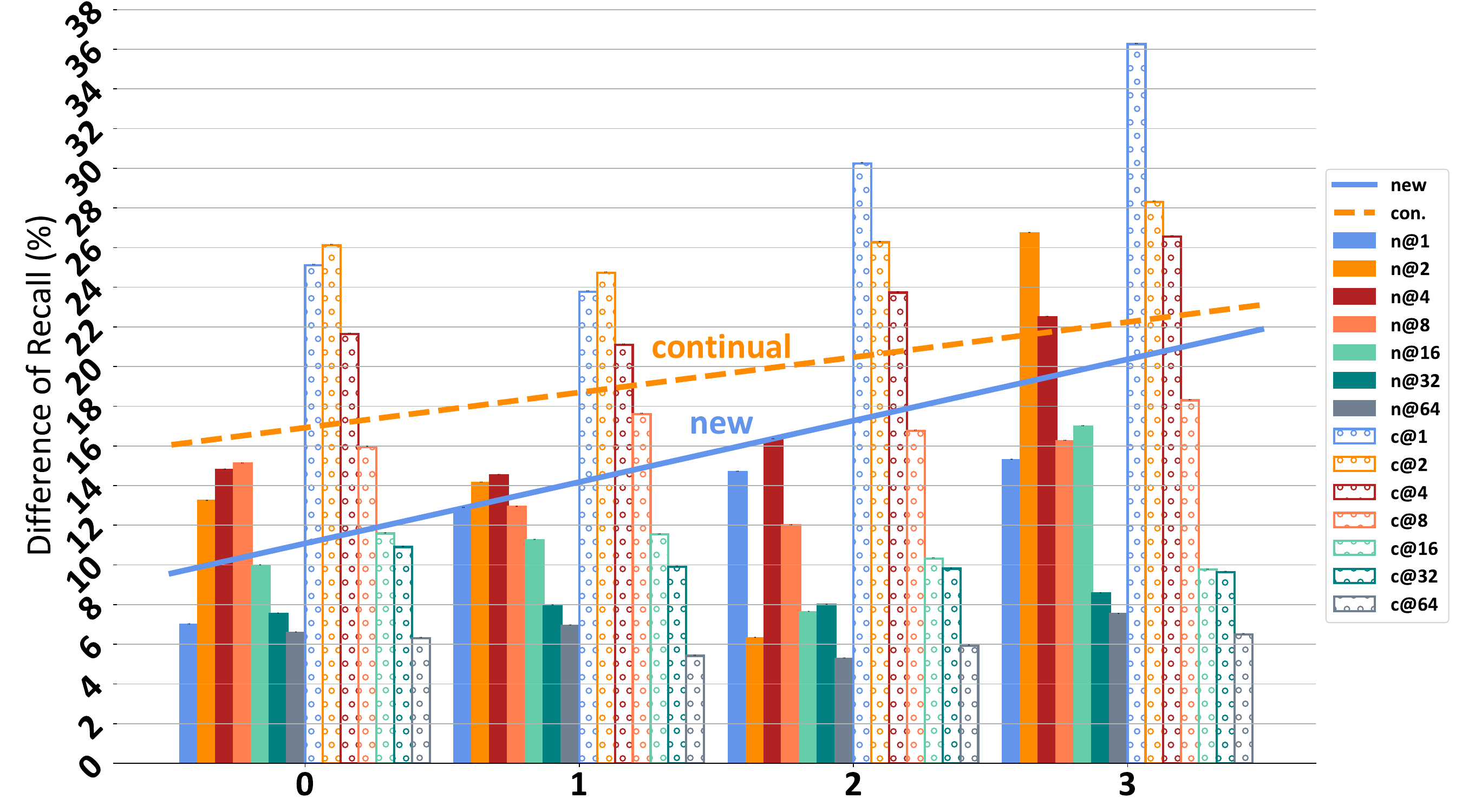}
\caption{This figure compares our model with the SpEL across 3-year gaps in \textit{only forward} settings. Improvements on the \textit{new} and \textit{continual entities} datasets are depicted by solid and hollow bars respectively. We observe a consistent growth in the performance gain for both types of entities (see regression lines) as the temporal gap increases, demonstrating the temporal robustness of our approach.
}\label{figure6}
\Description{figure6}
\end{figure}

Furthermore, our model's improvement margin diminishes as the recall metric threshold $N$ is increased, as illustrated in Figure~\ref{figure6}. The x-axis represents the year gap between training and testing datasets, while the y-axis shows our model's performance improvement over the SpEL model. Solid bars denote performance on the \textit{new entities} dataset and hollow bars on \textit{continual entities}. This diminishing effect is likely because at the @64 threshold, the model needs only to correctly predict one of the top 64 answers, thus allowing a higher error tolerance. Additionally, our model's performance enhancement grows with the year gap, as shown by the regression lines for both \textit{continual} and \textit{new} entities.

\subsection{Results on Low-degree Entities and Fairness Analysis}

\begin{figure}[t]
\centering
\includegraphics[width=0.93\linewidth]{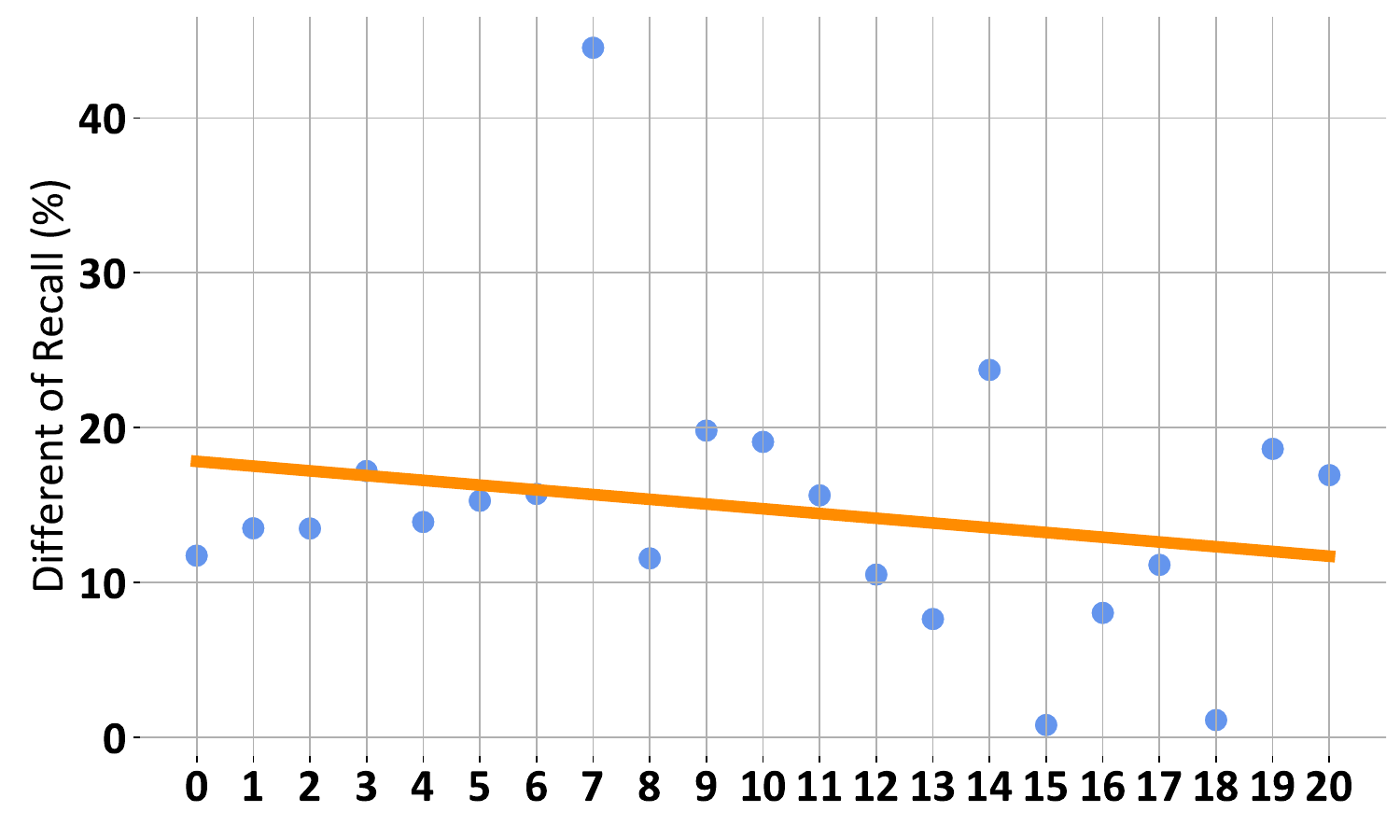}
\caption{CYCLE performance improvement over the BLINK model as the degree of target entities in the \textit{relation graph} increases (x-axis). The orange regression line illustrates a trend where CYCLE, on average, achieves better performance enhancements, particularly on low-degree entities.}\label{figure7}
\Description{figure7}
\end{figure}

In Figure~\ref{figure7}, we analyze the improvement of our model's performance on entities with different degrees. The model is trained on the 2019 \textit{new entities} training set and tested on the test sets from 2019 to 2022. The figure shows the entity's degree on the x-axis and the corresponding improvement in model performance compared to the BLINK model on the y-axis. We observe that our model has a more pronounced improvement for lower-degree entities. When facing entities with a relatively high degree and entities with more neighbors, our model still demonstrates an improvement, although it is less noticeable.

We hypothesize that this trend is partly due to sparse information of low-degree nodes. Such nodes are highly sensitive to new connections, making graph contrastive learning module especially effective at integrating new data and adapting to dynamic changes. This sensitivity is critical, as it allows low-degree nodes to more effectively capture and utilize new information, improving their embeddings and adaptability over time.

\subsection{Results on Non-temporal datasets}
Our model achieves state-of-the-art performance on temporally evolving~\ourdataset~dataset (see Table~\ref{table2}). In this section, we evaluate the performance of our model in a traditional setting, characterized by \textit{static} (i.e., lacking temporal evolution) EL datasets without graph-based input. Concretely, Table~\ref{table3} describes EL results using recall@$N$ metric, for our CYCLE model compared to BLINK and SpEL models on the ZESHEL and WikilinksNED \textit{static} datasets. Even without graph-based input and temporal evolution in these datasets, our model maintains consistent performance on par with BLINK, underscoring its adaptability. 

\begin{table}[t]
\caption{Performance of CYCLE on EL datasets that do not contain graph-based input and are static. 
Our model continues to exhibit competitive performance.
}
\label{table3}
\small
\setlength{\tabcolsep}{0.5mm}{
\begin{tabular}{@{}cccccccc@{}}
\toprule
                                                                                 \textbf{Dataset}  & \textbf{Model} & \textbf{@1} & \textbf{@4} & \textbf{@8} & \textbf{@16} & \textbf{@32} & \textbf{@64} \\ \midrule
\multirow{3}{*}{\textbf{\begin{tabular}[c]{@{}c@{}}Zeshel:\\ Forgotten Realms\end{tabular}}} & BLINK & 0.5183          & 0.7400          & 0.7950          & 0.8375          & 0.8683          & 0.8942          \\
                                                                                             & SpEL  & \textbf{0.5717} & \textbf{0.8092} & \textbf{0.8646} & \textbf{0.8969} & \textbf{0.9373} & \textbf{0.9498} \\
                                                                                             & CYCLE & 0.5150          & 0.7423          & 0.7963          & 0.8298          & 0.8676          & 0.8978          \\ \midrule
\multirow{3}{*}{\textbf{\begin{tabular}[c]{@{}c@{}}Zeshel: \\ Lego\end{tabular}}}            & BLINK & 0.4170          & 0.6647          & 0.7548          & 0.8090          & 0.8599          & 0.8841          \\
                                                                                             & SpEL  & \textbf{0.4672} & \textbf{0.7216} & \textbf{0.8113} & \textbf{0.8685} & \textbf{0.9197} & \textbf{0.9420} \\
                                                                                             & CYCLE & 0.4127          & 0.6726          & 0.7547          & 0.8059          & 0.8628          & 0.8825          \\ \midrule
\multirow{3}{*}{\textbf{\begin{tabular}[c]{@{}c@{}}Zeshel:\\ Star Trek\end{tabular}}}        & BLINK & 0.3717          & 0.5798          & 0.6475          & 0.7052          & 0.7563          & 0.7999          \\
                                                                                             & SpEL  & \textbf{0.4316} & \textbf{0.6358} & \textbf{0.7030} & \textbf{0.7574} & \textbf{0.8122} & \textbf{0.8534} \\
                                                                                             & CYCLE & 0.3749          & 0.5839          & 0.6493          & 0.7083          & 0.7528          & 0.7936          \\ \midrule
\multirow{3}{*}{\textbf{\begin{tabular}[c]{@{}c@{}}Zeshel:\\ Yugioh\end{tabular}}}           & BLINK & 0.2828          & 0.4769          & 0.5504          & 0.6094          & 0.6577          & 0.6935          \\
                                                                                             & SpEL  & \textbf{0.3361} & \textbf{0.5270} & \textbf{0.6056} & \textbf{0.6615} & \textbf{0.7110} & \textbf{0.7529} \\
                                                                                             & CYCLE & 0.2745          & 0.4764          & 0.5476          & 0.6075          & 0.6533          & 0.6946          \\ \midrule
\multirow{3}{*}{\textbf{WikilinksNED}}                                                       & BLINK & 0.1721          & 0.4192          & 0.5467          & 0.6505          & 0.7340          & 0.7907          \\
                                                                                             & SpEL  & \textbf{0.2315} & \textbf{0.4761} & \textbf{0.5976} & \textbf{0.7084} & \textbf{0.7913} & \textbf{0.8414} \\
                                                                                             & CYCLE & 0.1746          & 0.4276          & 0.5657          & 0.6584          & 0.7275          & 0.7934          \\ \midrule
\multirow{3}{*}{\textbf{Average}}                                                            & BLINK & 0.3524          & 0.5761          & 0.6589          & 0.7223          & 0.7752          & 0.8125          \\
                                                                                             & SpEL  & \textbf{0.4076} & \textbf{0.6339} & \textbf{0.7164} & \textbf{0.7785} & \textbf{0.8343} & \textbf{0.8679} \\
                                                                                             & CYCLE & 0.3503          & 0.5806          & 0.6627          & 0.7220          & 0.7728          & 0.8124          \\ \bottomrule
\end{tabular}}
\end{table}

\section{Conclusion and Future Work}

This paper introduces \textbf{CYCLE}, a model specifically designed to use graph contrastive learning to mitigate temporal degradation in evolving Knowledge Graphs (KGs). The model captures cross-year changes between entities by utilizing newly added or removed graph nodes as positive and negative samples in a contrastive learning framework. We conducted experiments on three Entity Linking (EL) datasets, setting a new benchmark on the temporally evolving~\ourdataset~dataset. Specifically, our model achieved a 13.90\% performance improvement over the SpEL model when the time gap is one year, and this improvement increased to 17.79\% as the gap extended to three years. Additionally, our model demonstrated competitive performance on static datasets. Looking ahead, we identify two key areas for future research:

\textbf{Multimodal Temporal KGs with Contrastive Learning.} In addition to temporal information, TKGs can also contain multimodal data.
This multimodal data can be used further to improve the performance of contrastive learning for TKGs. For example, a pair of entities with a relationship in multiple timestamps that are also mentioned in the audio can be considered a more informative positive sample than a pair of entities with a relationship in multiple timestamps but not in any other type of data.

\textbf{Temporal KGs with Large Language Models.} Considering the challenge of aligning similar concepts across languages where direct translations often fail to convey identical meanings, using Large Language Models (LLMs) for aligning conceptually similar entities in multilingual KGs
can improve the coherence of these graphs. If entity descriptions and entity relationships exist at language $A$, the model could benefit from these diverse language resources.

\begin{acks}
The first author is supported by the China Scholarship Council (NO. 202206540007) and the University of Amsterdam. This funding source had no influence on the study design, data collection, analysis, or manuscript preparation and approval. This work is partially supported by the EU’s Horizon Europe programme, in the ENEXA project (grant Agreement no. 101070305).
Co-funded by the European Union (grant Agreement no. 101146515). Views and opinions expressed are however those of the authors only and do not necessarily reflect those of the European Union or European Research Executive Agency. Neither the European Union nor the granting authority can be held responsible for them. 
\end{acks}

\bibliographystyle{ACM-Reference-Format}
\balance
\bibliography{main}

\end{document}